\documentclass[runningheads]{llncs}
\usepackage{graphicx}
\usepackage{amsmath,amssymb} 
\usepackage{color}
\usepackage{url}
\usepackage{verbatim}
\usepackage{graphicx}
\usepackage{booktabs}
\usepackage{multirow}
\usepackage{cite}
\usepackage{algorithmic}
\usepackage{algorithm}
\usepackage{array}
\usepackage{subfigure}
\usepackage{textcomp}
\usepackage{stfloats}
\usepackage{bm}
\begin{document}
\title{DFA: Dynamic Feature Aggregation for Efficient Video Object Detection}
\titlerunning{DFA}
%
\author{Yiming Cui}
%
\authorrunning{Yiming Cui et al.}
%
\institute{University of Florida}
%
\maketitle              
\begin{abstract}
Video object detection is a fundamental yet challenging task in computer vision. One practical solution is to take advantage of temporal information from the video and apply feature aggregation to enhance the object features in each frame. Though effective, those existing methods always suffer from low inference speeds because they use a fixed number of frames for feature aggregation regardless of the input frame. Therefore, this paper aims to improve the inference speed of the current feature aggregation-based video object detectors while maintaining their performance. To achieve this goal, we propose a vanilla dynamic aggregation module that adaptively selects the frames for feature enhancement. Then, we extend the vanilla dynamic aggregation module to a more effective and reconfigurable deformable version. Finally, we introduce inplace distillation loss to improve the representations of objects aggregated with fewer frames. Extensive experimental results validate the effectiveness and efficiency of our proposed methods: On the ImageNet VID benchmark, integrated with our proposed methods, FGFA and SELSA can improve the inference speed by $31\%$ and $76\%$ respectively while getting comparable performance on accuracy. 
\end{abstract}
%
%
%
\section{Introduction}
Object detection is an essential task in computer vision which aims to localize and categorize objects of interest in a single or sequence of images \cite{sun2021sparse, he2017mask, ren2015faster, girshick2015fast, cui2021tf, chen2020memory, zhu2017flow, wu2019sequence, liu2020large}. With the excellent performance of deep learning-based computer vision methods on image object detection tasks \cite{sun2021sparse, he2017mask, ren2015faster, girshick2015fast}, researchers have begun to extend image object detection to the more challenging video domain. Compared with still images, videos have the issues of feature degradation caused by camera jitters or fast motion that rarely happen in the image domains \cite{zhu2017deep,chen2020memory}, which increase the difficulty of object detection in videos. Therefore, directly applying object detectors from image domains on a frame-by-frame basis for video analysis always produces poor performance. Existing works can be divided into two directions to solve the issues caused by video feature degradation. 

Since the same object always reappears in multiple frames, videos can provide rich temporal information, which provides hints for video analysis \cite{yan2022gl, Cui_2022_WACV, liu2021sg}. Therefore, one direction for video object detection is to exploit this temporal information with post-processing pipelines \cite{han2016seq, Kang_2016, kang2017t} where still image object detection approaches are first applied on each frame. Then those detected objects are assembled across frames with temporal hints like motion estimation or object tracking. However, those approaches are not trained end-to-end, and the detection results on a single frame and across frames cannot be optimized jointly. Therefore, if the predictions in a single frame are inaccurate, they cannot be optimized and refined during the post-processing procedure.

Another direction for video object detection is aggregating features across multiple frames to eliminate feature degradation in videos \cite{liu2020video, liu2021densernet, yan2021hierarchical, liu2020indoor, liu2021visual, cui2021geometric}. These methods assume that frames with poor features only account for a small ratio compared with the whole video sequences. By aggregating temporal features, the performance of video object detection can be boosted. These methods can be categorized as local, global, and combinations, depending on how to aggregate features. The first sub-direction methods \cite{zhu2017deep, zhu2017flow, wang2018fully} exploit the local temporal information in videos to enhance the target frame features in a short time range and ignore the global information. To address this issue, the second sub-direction methods \cite{deng2019relation, han2020exploiting, shvets2019leveraging, han2021class} introduce attention modules to use global temporal information. However, these methods ignore local temporal information due to GPU memory limits or computational constraints. The third sub-direction methods \cite{chen2020memory, jiang2020learning} make a combination of local and global temporal information but always suffer from low inference speeds. 

\begin{figure}[!tb]
    \centering
    \includegraphics[width=11cm]{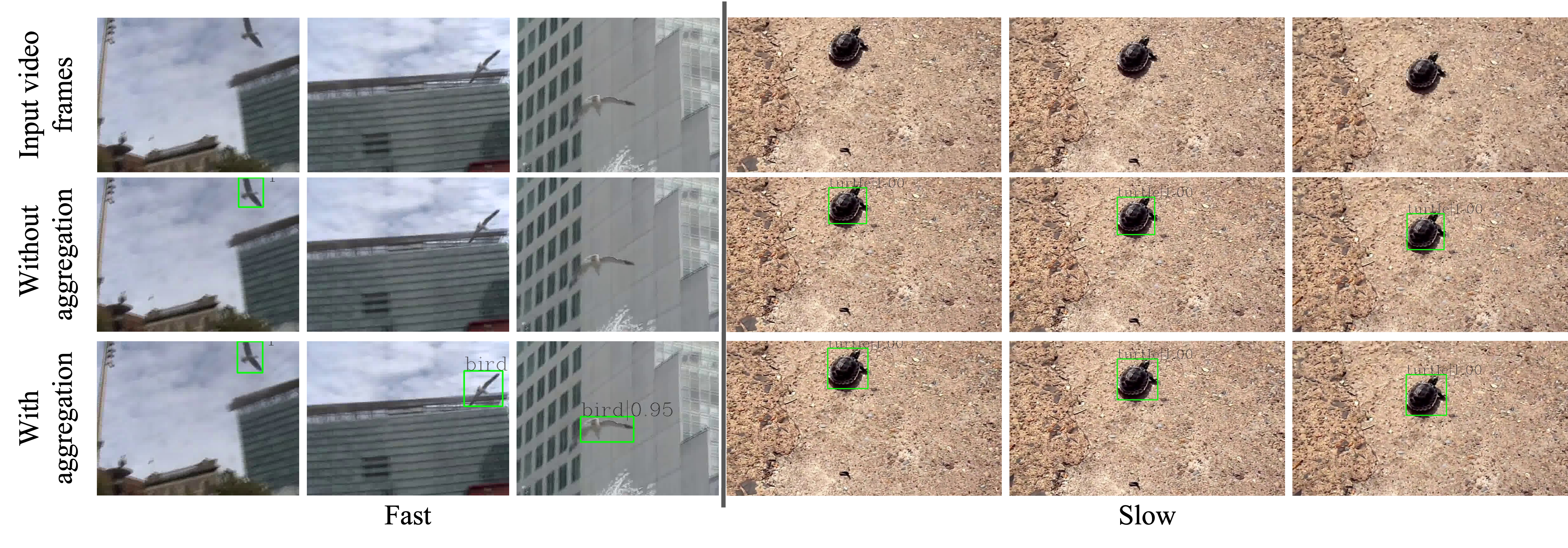}
    \caption{Comparison of video object detection results with/without feature aggregation on objects with different motion speeds.} 
    \label{fig: videoDegrad}
\end{figure}

Though getting better performance than post-processing methods, those feature aggregation-based object detectors always have a lower inference speed. Therefore, besides focusing on the performance of video object detection, recent works \cite{jiang2020learning,Xu2020CenterNetHP, chen2018optimizing, yao2020video, jiang2019video} also design efficient frameworks to improve the inference speeds. However, these methods are designed for a specific framework and cannot be generalized to other video object detectors. What makes things worse, these approaches are always efficient during the inference process at the sacrifice of performance like accuracy or recall.

Unlike the existing efficient video object detectors, we aim to design a plug-and-play module that can be integrated into most existing methods to balance their inference speeds and performances. To achieve this, we first notice that the low inference speeds of the current feature aggregation-based object detectors are caused by their aggregation processes, which are proportional to the number of frames used for aggregation \cite{zhu2017flow, zhu2017deep,cui2021tf, chen2020memory}. It is natural to think whether it is necessary to always use a fixed number of frames for feature aggregation. For objects with fast motion speeds \footnote{For better analysis, we use the same way as FGFA \cite{zhu2017flow} to categorize objects in every single frame based on their motion speeds.}, feature aggregation can improve the video object detection performance. As shown in Figure \ref{fig: videoDegrad}(a), the flying bird cannot be correctly detected in multiple frames without feature aggregation. On the contrary, when the objects are with slow motion speeds, as shown in Figure \ref{fig: videoDegrad} (b), original Faster R-CNN \cite{ren2015faster} without any feature aggregation can already detect the turtle in the current frame correctly. Therefore, using too many aggregation frames for videos with slow motion is unnecessary since the model with a few aggregation frames or even without aggregation can already perform well.

In this paper, we attempt to improve the efficiency of the current feature aggregation-based video object detectors in a simple yet effective way. We notice that there is no need to always use a fixed number of frames to aggregate features for video object detection regardless of the inputs. Therefore, we design modules to aggregate features dynamically. We first propose a vanilla dynamic feature aggregation strategy which can adaptively select frames for aggregation based on the inputs. Then, we extend the vanilla strategy to a deformable version which is more effective and reconfigurable. Finally, we introduce an inplace distillation loss to enhance the object feature representations when only a few frames are used for aggregation. Our contributions can be summarized as follows:

\begin{itemize}
    \item To the best of our knowledge, we are the first to adaptively and dynamically aggregate features for video object detection to balance the model efficiency and performance.
    \item We design a vanilla dynamic feature aggregation (DFA) module and then extend it to a deformable version which can adaptively and reconfigurably enhance the object feature representations based on the input frame. Inplace distillation loss is introduced to improve the feature representations of those aggregated with fewer frames for better performance.
    \item Our proposed method is a plug-and-play module which can be integrated into most of the recent state-of-the-art video object detectors. Evaluated on the ImageNet VID benchmark, the performance of video object detection can be preserved with a much better inference speed when integrated with our proposed method.
\end{itemize}

\section{Related Works}
\noindent\subsubsection{Still image object detection.} Image object detection task aims to localize and categorize objects of interest in a still image. Current deep learning-based models can be classified into two main directions: Two-stage object detector and one-stage object detector. Among them, R-CNN-based two-stage object detectors \cite{ren2015faster, girshick2015fast, he2017mask, lin2017feature, Cai_2019} first generate a fixed number of proposals with Region Proposal Network (RPN) \cite{girshick2015fast} to localize and classify the object candidates coarsely. Then, they refine these proposals to output fine-grained predictions. To improve the inference speed of those models mentioned above, one-stage models \cite{law2018cornernet, liu2016ssd,redmon2016you, tian2019fcos} are introduced to predict the locations and categories of objects directly based on the extracted features from CNN without region proposals. For simplicity and generalization, our method is built based on Faster R-CNN \cite{ren2015faster}, which is one of the state-of-the-art object detectors and can be easily extended to others.

\noindent\subsubsection{Video object detection.} Different from image object detection, the video object detection task must handle situations caused by motion to generate good predictions in each frame. Post-processing-based methods detect every frame separately and assemble those detected objects with various metrics like optical flow. Seq-NMS \cite{han2016seq} assembles bounding boxes at different frames with the criteria of IoU threshold and re-ranks the linked bounding boxes. TCN \cite{Kang_2016} uses tubelet modules and applies a temporal convolutional network to embed temporal information to improve the detection across frames. Despite the simplicity, those methods are not trained end-to-end and perform poorly.

To solve the issues, feature aggregation-based methods \cite{zhu2017deep, zhu2017flow, chen2020memory, wu2019sequence} usually enhance the object representations using the temporal information to eliminate the feature degradation caused by motions. Among them, FGFA \cite{zhu2017flow}  first warps the feature maps from the local adjacency frames to the keyframe based on the flow motion and then aggregates those warped features to improve the object representations for the following detection network. SELSA \cite{wu2019sequence} aggregates features in a global full-sequence level. In SELSA, proposals across space-time domains with similar semantics are linked, and their features are weight-averaged for aggregation to provide richer information to handle issues like motion blur and pose changes. MANet\cite{wang2018fully} jointly aggregates object features on both pixel-level and instance-level. The pixel-level aggregation is used to model detailed motion, while the instance-level calibration is introduced to capture global motion cues. MEGA \cite{chen2020memory} takes global and local information into account where global features are first aggregated into local features. Then these global-enhanced local features are fused into the key frame for better detection performance. TF-Blender \cite{cui2021tf} improves the feature aggregation process using the temporal relations between frames. TransVOD \cite{he2021end} introduces the Transformer to aggregate the spatial and temporal information in a multi-head self-attention mode. 

Compared with post-processing-based methods, feature aggregation-based video object detectors usually perform better with a lower inference speed. In this paper, we mainly focus on feature aggregation-based methods. 

\begin{figure} [!tb]
    \centering
    \subfigure[Existing methods]{
    \includegraphics[height=3.2cm]{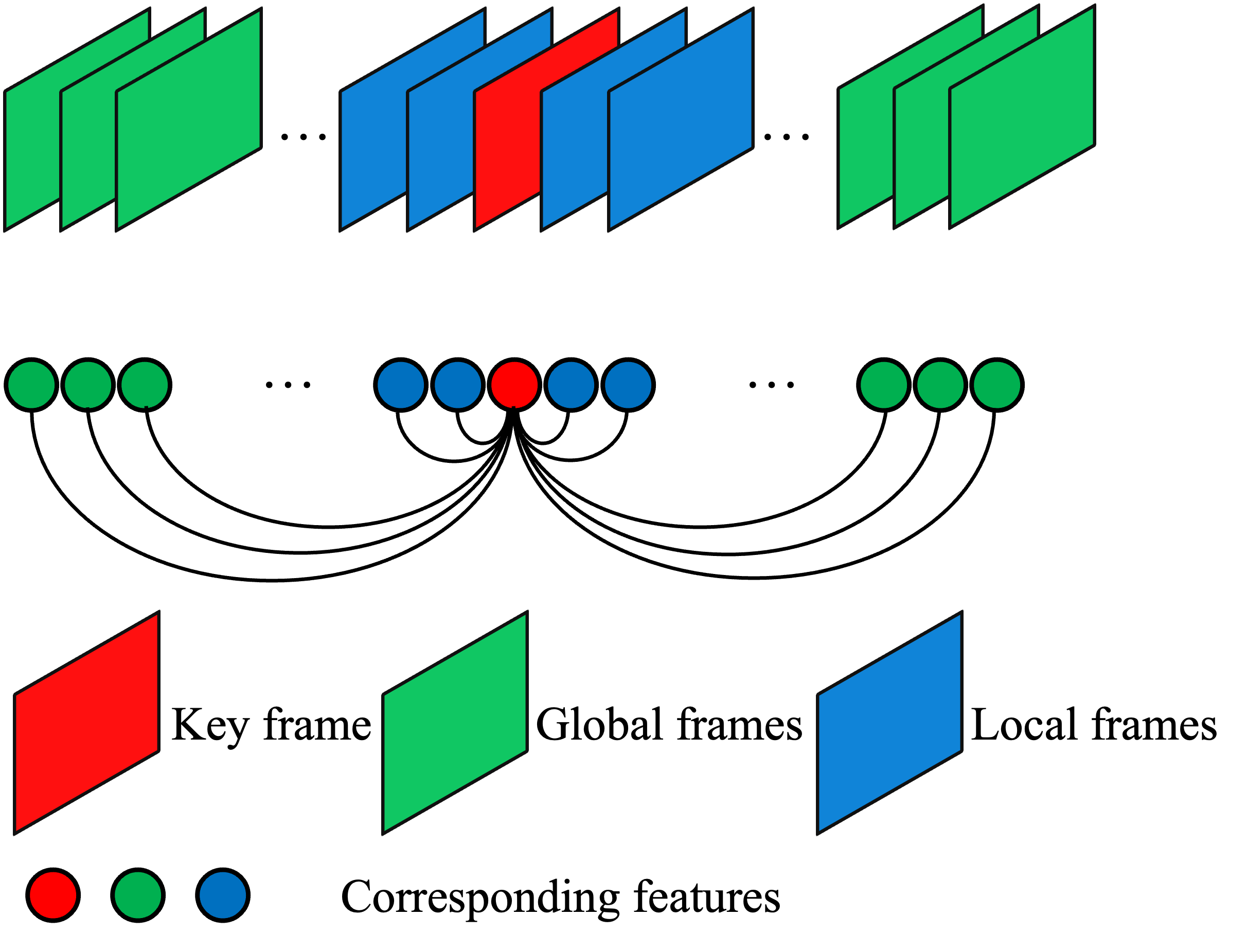}
    }
    \subfigure[Our method]{
    \includegraphics[height=3.2cm]{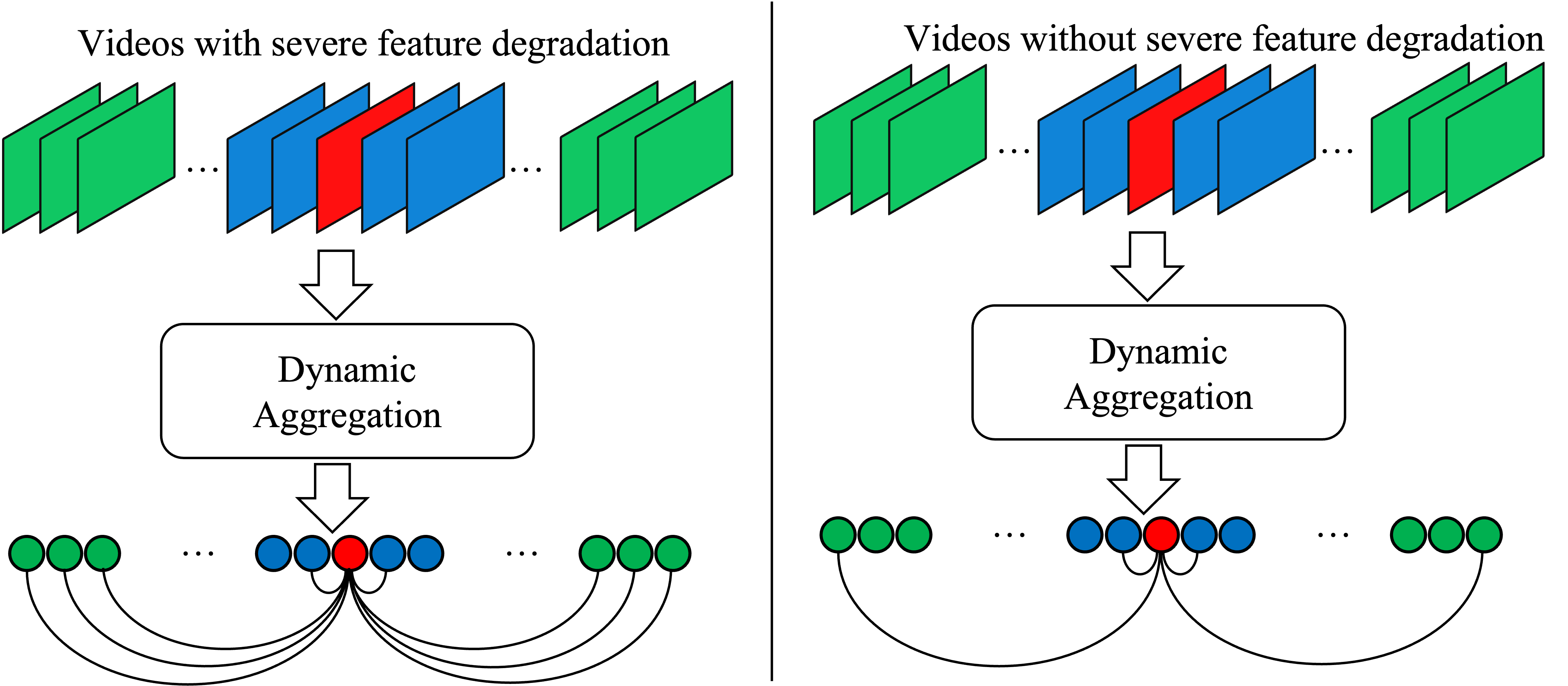}
    }
    \caption{The framework of our proposed method, which uses dynamic aggregation strategies for video object detection tasks. (a) Current methods aggregate features with a fixed number of frames (e.g., $10$) regardless of the input frames. (b) Our methods adaptively select frames for dynamic aggregation according to the input frame. For videos with severe feature degradation, $8$ frames are selected to enhance the key frame representation. For videos with high qualities, $4$ frames are used for aggregation for fast inference speed.}
    \label{fig:fixAgg}
\end{figure}

\noindent\subsubsection{Efficient networks.}  Though deep learning-based methods perform better on multiple computer vision tasks, their complexities become higher, making them unsuitable for applications with constrained computational budgets but a short response time like mobile platforms. Therefore, recent works \cite{jiang2020learning, howard2017mobilenets, sandler2018mobilenetv2, Xu2020CenterNetHP, tan2019efficientnet} begin to focus on how to speed up the detection process for real-time applications. Towards this goal, lightweight networks like Mobile-Net \cite{howard2017mobilenets, sandler2018mobilenetv2}, Efficient-Net \cite{tan2019efficientnet} and automated neural architecture search models \cite{liu2018progressive,wu2019fbnet} are introduced to take the place of heavy backbones like ResNet \cite{he2016deep} to reduce the computation complexity for mobile applications. Besides replacing the backbones, LSTS \cite{jiang2020learning} learns semantic-level spatial correspondences between neighboring frames to reduce the information redundancy in video frames to accelerate the detection process. CenterNet-HP \cite{Xu2020CenterNetHP} replaces two-stage detectors like Faster R-CNN \cite{ren2015faster} with one-stage model CenterNet \cite{duan2019centernet} for real-time video object detection. Detection results from previous frames are propagated in the form of a heatmap to enhance the performance of the future frames. Other works \cite{wang2021real, wang2019fast} improve the detection speeds with the help of compressed video information. Though efficient during inference, those methods usually require carefully designed modules and make great changes to the existing video object detectors, making it infeasible to generalize to other methods. Also, these methods generally have a worse performance despite high inference speeds. On the contrary, our proposed approaches are plug-and-play modules which can be easily integrated into the existing detectors to balance their efficiency and performance.

Recently, dynamic networks have been introduced, which allow selective inference paths. Slimmable networks \cite{yu2018slimmable, yu2019autoslim, yu2019universally} are models trained executable at different widths, which can be adaptive to multiple computational resources and get even better performance compared with their counterparts trained individually. For object detection, dynamic proposals are introduced for efficient inferernce \cite{cui2022dynamic}. In this paper, we borrow the idea from slimmable networks to make our model adaptive to different input videos and able to adjust the numbers of frames for aggregation according to the input frame for video object detection.
\begin{figure}[!tb]
    \centering
    \includegraphics[width=10cm]{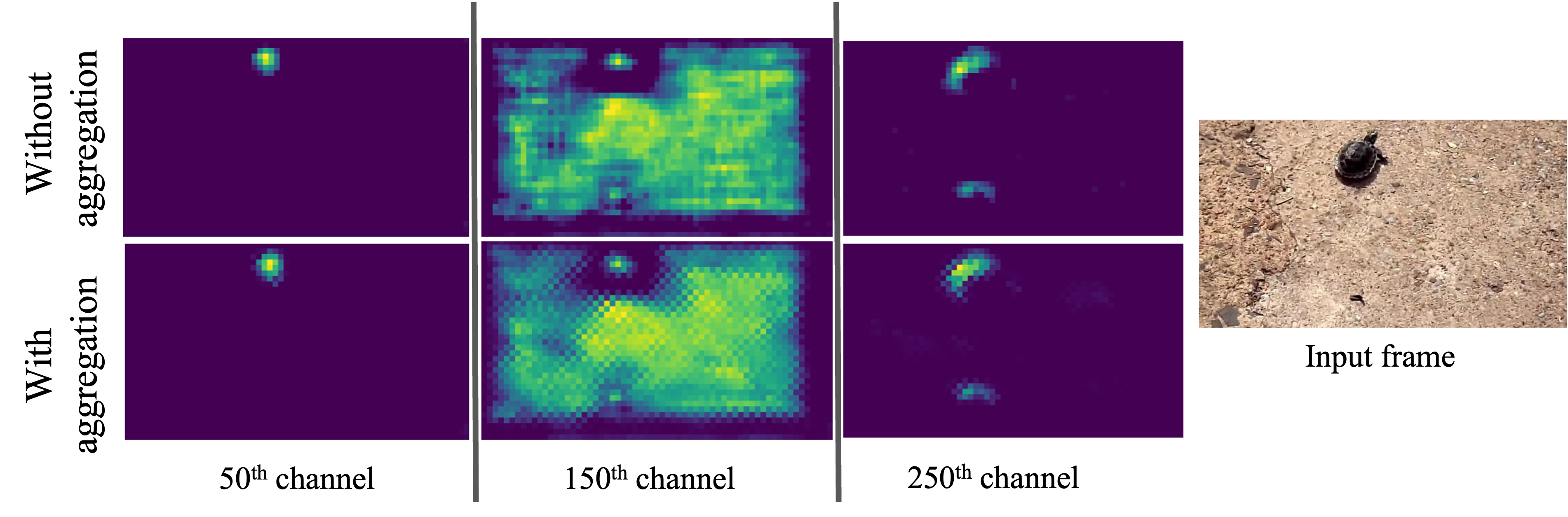}
    \caption{Comparison of features maps with/without aggregation on objects with slow motion speeds.}
    \label{fig:aggVSnonAggFeatureMap}
\end{figure}
\section{Methodology}
The key idea of our method is to replace the fixed number of frames with a dynamic size in the current feature aggregation-based video object detectors. Therefore, instead of using fixed frames, our model can adaptively choose the frames for aggregation according to the inputs, as shown in Figure \ref{fig:fixAgg}. Our proposed method is a plug-and-play module which can be easily integrated into most feature aggregation-based video object detectors. In the following sections, we first review the current feature aggregation-based video object detectors and analyze the inefficiency of their aggregation processes. Then, we propose a vanilla dynamic aggregation strategy to adapt the model to different input frames. Next, we extend the vanilla dynamic aggregation to a deformable version which is more effective and reconfigurable. Finally, inplace distillation loss is introduced to enhance the feature representations aggregated with fewer adjacent frames.

\subsection{Preliminary}
The canonical feature aggregation methods \cite{zhu2017flow,wu2019sequence,Wang_2018_ECCV,chen2020memory, cui2021tf, deng2019relation} generally work on a fixed number of frames $k$, which can be summarized as: Given the current frame $\bm{I}_i$ and its neighboring frames $\bm{I}_{j} \in  \mathcal{N}\left(\bm{I}_i\right)$, their corresponding features $\bm{f}_{j}$ are weighted averaged in order to aggregate the temporal feature $\Delta{\bm{f}}_{i}$:
\begin{equation}
\begin{aligned}
      \Delta \bm{f}_i &=
      \sum_{\bm{I}_j \in \mathcal{N}\left(\bm{I}_i\right) }\left({w}_{ij}\times{\bm{f}}_{j}\right),
    \end{aligned}
    \label{eq: aggregation}
\end{equation}
where $w_{ij}$ denotes the weights for aggregation and the size of $\mathcal{N}\left(\bm{I}_i\right)$ is $k$. Then the aggregated feature $\Delta \bm{f}_i$ is fed into a task network $\mathcal{N}_{task}$ for object detection:
\begin{equation}
    \bm{b}_i, \bm{c}_i = \mathcal{N}_{task} \left(\Delta \bm{f}_i\right),
    \label{eq: task}
\end{equation}
where $\bm{b}_i, \bm{c}_i$ denote the predicted bounding boxes and their corresponding categories in the current frame $\bm{I}_i$. However, the performance and inference speed of these feature aggregation-based models are heavily influenced by $k$. For example, when $k$ decreases from $31$ to $3$, the inference speed of FGFA \cite{zhu2017flow} can increase from $5.8$ FPS to $20.6$ FPS (tested on a single Titan RTX GPU) while the mAP drops from $74.6\%$ to $72.5\%$.

Therefore, we wonder whether it is necessary to use so many frames (e.g., 21 frames) for feature aggregation regardless of the input videos. Figure \ref{fig:aggVSnonAggFeatureMap} compares the feature maps with and without feature aggregation on objects with slow motion speeds. We visualize the $50$-th, $150$-th and $250$-th feature channels in Figure \ref{fig:aggVSnonAggFeatureMap}. As shown in the figure, there is little difference between the feature maps with and without aggregation. Meanwhile, we also calculate the cosine similarity scores of the whole feature map with and without aggregation, which is $0.9819$. Therefore, for objects with slow motion speeds, there is not too much improvement with feature aggregation.   
\subsection{Vanilla Dynamic Aggregation}
In the current feature aggregation based video object detectors \cite{zhu2017flow, wu2019sequence, gong2021temporal, wang2018fully}, as described in Equation \ref{eq: aggregation}, $w_{ij}$ is calculated as the cosine similarity between the neighboring feature $\bm{f}_j$ and $\bm{f}_i$, which is unrelated to $k$. Since $k$ will not affect the aggregation process during inference time, it is possible to update the feature aggregation module in the current methods to a dynamic version, where an adaptive number of frames is applied to eliminate feature degradation to boost the performance of video object detection. 
 
A simple idea to achieve this is to classify the current frames into multiple categories and determine the number of frames used for aggregation based on the categories. That is how our vanilla dynamic aggregation method comes out. In detail, we make the number of frames for feature aggregation dynamic based on the input frame: For frames where objects are with low motion speeds, fewer or even no neighboring frames are taken into account for feature aggregation. On the contrary, for those frames with severe feature degradation, more or even the whole neighboring frames are used to boost the detection performance.

To make the number of frames for feature aggregation dynamic, instead of a fixed number $k$,  we use $k_v$ frames for aggregation, which is determined by the current input frame $\bm{I}_i$. To achieve this, we first categorize the current frame $\bm{I}_i$ into $\theta$ categories based on the motion speeds of the objects in $\bm{I}_i$, where $\theta$ is a configurable parameter. Then, we use a function $\mathcal{S}_v\left(k, \delta\right)$ to determine $k_v$, where $\delta$ is an integer within the range of $\left(0, \theta\right]$ to represent the category of $\bm{I}_i$. Then, the formulation of $\mathcal{S}_v\left(k,\delta\right)$ can be represented as:  
\begin{equation}
    k_v = \mathcal{S}_v \left(k, \delta\right) = \left\lceil\delta \frac{k}{\theta}\right\rceil, 
    \label{eq: vanillaDet}
\end{equation}
where $\lceil\rceil$ denotes the ceiling function. When $\theta$ is defined, $k_v$ will have multiple discrete choices depending on $\delta$. For example, when $\theta$ is chosen to be $3$, $\delta=1$ represents frames where objects are with slow motion speeds, and $k_v=\lceil\frac{k}{3}\rceil$ frames are used for feature aggregation for fast inference speed. Similarly, $\delta=3$ means the current frame contains objects which move fast and we use $k_v=k$ frames to eliminate the feature degradation for a better performance. Given the number of frames $k_v$ for the current frame $\bm{I}_i$, we slice a subset of the neighboring $\mathcal{N}_v\left(\bm{I}_i\right)$ from the whole range $\mathcal{N}\left(\bm{I}_i\right)$ using:
\begin{equation}
    \mathcal{N}_v\left(\bm{I}_i\right) = \mathcal{G}\left(\mathcal{N}\left(\bm{I}_i\right), k_v\right),
    \label{eq: vanillaSample}
\end{equation}
where $\mathcal{G}\left(\cdot, k_v\right)$ is a sampling function to select $k_v$ neighboring frames from $k$ total neighboring frames.  Therefore, Equation \ref{eq: aggregation} will be updated as: 
\begin{equation}
      \Delta \bm{f}_i^v = \sum_{\bm{I}_j \in \mathcal{N}_v\left(\bm{I}_i\right)} \left({w}_{ij}^v\times{\bm{f}}_{j}\right), \\
    \label{eq: vanilla}
\end{equation}
where $\Delta \bm{f}_i^v, w_{ij}^v$ represent the aggregated features of frame $\bm{I}_i$ with dynamic neighborhood $\mathcal{N}_v\left(\bm{I}_i\right)$ and the corresponding weights. 

During the training and inference processes, the category of the current frame $\bm{I}_i$, denoted as $\delta$, is predicted based on the features of $\bm{I}_i$ and its neighboring frames $\mathcal{N}\left(\bm{I}_i\right)$. In detail, $\bm{f}_i$ is first concatenated with $\bm{f}_j$ and then fed into a mini-network $\mathcal{N}_{mot}^v$ to predict the category of $\bm{I}_i$, summarized as:
\begin{equation} 
    \delta = \mathcal{N}_{mot}^v \left(\texttt{cat}\left({\bm{f}_i, \bm{f}_j}\right)\right), \qquad\forall \bm{I}_j \in \mathcal{N}\left(\bm{I}_i\right)
    \label{eq: motionEst}
\end{equation} 

Following FGFA \cite{zhu2017flow} and MEGA \cite{chen2020memory}, we measure the motion speed of an object in a frame with motion IoU, denoted as $s_m$, using the averaged intersection-over-union (IoU) scores with its corresponding instances in the neighboring frames. Then, we divide each frame into $\theta$ classes based on $s_m$ to generate the ground truth category $\delta^{gt}$. For example, when $\theta$ is set to be $3$, objects are classified into slow ($s_m > 0.9$), medium ($s_m \in \left[0.7, 0.9\right]$) and fast ($s_m < 0.7$) groups, respectively. Therefore, each frame is divided into $3$ categories based on the motion speeds of the objects it contains. Cross entropy loss ($\mathcal{L}_{CE}$) between $\delta$ and $\delta^{gt}$ is calculated as the loss $\mathcal{L}_{mot}^v$ to optimize the network $\mathcal{N}_{mot}^v$, as:
\begin{equation}
    \mathcal{L}_{mot}^v = \mathcal{L}_{\text{CE}}\left(\delta, \delta^{gt}\right)
    \label{eq: vanillaMot}
\end{equation}
\subsection{Deformable Dynamic Aggregation}
With our proposed vanilla dynamic aggregation method, the current video object detectors' performance and inference speed can theoretically be well balanced. However, there are two issues. 

The categories to determine $k_v$ need to be predefined during the training process and are not reconfigurable at inference time. In other words, a well-trained model is not adaptive to multiple configurations. Take $\theta=3$, which represents objects with slow, medium, and fast motion speeds, as an example. Given a frame with the motion IoU of $s_m=0.75$, it is always categorized as medium motion speeds, which requires $\lceil\frac{2}{3}k\rceil$ frames for aggregation. If we would like to regard the frame as slow motion speeds to use fewer frames for aggregation when computational resources are limited, we need to modify the category ranges (for example, $s_m \in \left[0.5, 0.7\right]$ for the medium group) and train a new model again. It is inconvenient and unsuitable for real-world applications when $\theta$ is set to be very large or the category ranges are switched frequently.  

Moreover, experiments show that the vanilla dynamic aggregation module does not perform well in detecting objects of small sizes. Most of the existing feature aggregation-based video object detectors use Faster R-CNN \cite{ren2015faster} as the baseline method without feature pyramid networks \cite{lin2017feature}. Therefore, more frames are required to enhance the feature representations of small objects during the aggregation process. However, in the vanilla dynamic aggregation module, the sizes of objects in the frames are not considered.

To solve the issues mentioned above, we extend the vanilla dynamic aggregation module to a deformable version, which is more effective and reconfigurable. Instead of classifying the input frame $\bm{I}_i$ into $\theta$ categories, we use a function $\sigma$ to project the $s \in \left[0, 1\right]$ in the range of $0$ and $1$, where $s$ is a score which takes both the motion IoU $s_m \in\left[0, 1\right]$ and size $s_s \in \left[0, 1\right]$ of objects in the current frame $\bm{I}_i$ into account. Therefore, Equation \ref{eq: vanillaDet} is updated to be:
\begin{equation}
    k_d = \mathcal{S}_d \left(k, s\right) = \left\lceil\sigma\left(s\right) k\right\rceil = \left\lceil\sigma\left(s_m s_s\right) k\right\rceil
    \label{eq: deformDet}
\end{equation}

During the inference time, we can determine $k_d$ by selecting $\sigma$ in configure files. In real-world applications, when there are enough computational resources like applications on servers, we can choose $\sigma$, which casts $s\in[0,1]$ in the range of $0$ and $1$. When there are not enough resources, like applications on cellphones or servers where partial machines are under maintenance, we can reload the configure file where a new $\sigma$ projects $s$ in a new range (e.g. $\left[0, 0.5\right]$) to use fewer frames for aggregation without the need of training a new model. 

Similarly, given $k_d$ for the current frame $\bm{I}_i$, the sampled neighboring frames $\mathcal{N}_d\left(\bm{I}_i\right)$ is represented as Equation \ref{eq: deformSample} and deformable dynamic aggregation process is summarized as Equation \ref{eq: deform}.
\begin{equation}
    \mathcal{N}_d\left(I_i\right) = \mathcal{G}\left(\mathcal{N}\left(I_i\right), k_d\right),
    \label{eq: deformSample}
\end{equation}
\begin{equation}
      \Delta \bm{f}_i^d = \sum_{\bm{I}_j \in \mathcal{N}_d\left(\bm{I}_i\right) }({w}_{ij}^d\times{\bm{f}}_{j}), \\
    \label{eq: deform}
\end{equation}
where $\Delta f_i^d, w_{ij}^d$ represents the enhanced features of frame $\bm{I}_i$ with dynamic neighborhood $\mathcal{N}_d\left(\bm{I}_i\right)$ generated from the deformable dynamic aggregation module and the corresponding weights.

During the training and inference processes, we use mini-networks $\mathcal{N}_{mot}^d$ and $\mathcal{N}_{size}$ to estimate the averaged motion IoU $s_m$ and size $s_s$ of objects in the current frame $I_i$, respectively. Similar to Equation \ref{eq: motionEst}, the above process can be represented as:
\begin{equation}
\begin{aligned}
  s_m &= \mathcal{N}_{mot}^d \left(\texttt{cat}\left({\bm{f}_i, \bm{f}_j}\right)\right), \qquad\forall \bm{I}_j \in \mathcal{N}\left(\bm{I}_i\right) \\
  s_s &= \mathcal{N}_{size}\left(\bm{f}_i\right)
\end{aligned}
\end{equation}

For each frame $\bm{I}_i$, we calculate the averaged bounding box area of the objects it contains as the ground truth $s_s^{gt}$ for $\mathcal{N}_{size}$. Motion IoU ground truths $s_m^{gt}$ are measured with the same pipeline as FGFA \cite{zhu2017flow} and the vanilla dynamic aggregation module. Then, mean square error loss $\mathcal{L}_\text{{MSE}}$ is applied to optimize $\mathcal{N}_{mot}^d$ and $\mathcal{N}_{size}$ as:
\begin{equation}
\begin{aligned}
    \mathcal{L}^d_{mot} &= \mathcal{L}_\text{{MSE}}\left(s_m, s_m^{gt}\right) \\
    \mathcal{L}_{size} &= \mathcal{L}_\text{{MSE}}\left(s_s, s_s^{gt}\right)
\end{aligned}
\end{equation}

\subsection{Inplace Distillation Loss}
Our proposed method aims to balance the inference speed and performance of the existing feature aggregation-based video object detectors. Therefore, when it comes to the situation that $k_v\left(k_d\right)$ is small, we would like features $\Delta \bm{f}_i^v (\Delta \bm{f}_i^d)$ aggregated with $k_v(k_d)$ frames similar to $\Delta \bm{f}_i$ aggregated with $k$ frames. Here we borrow the idea from knowledge distillation that the performance of student models can be boosted when trained with the soft predictions of teacher models.

In our method, we treat the full model with $k$ frames for aggregation as the teacher network and those with fewer frames $k_v(k_d) < k$ as the student models. We add an extra $\mathcal{L}_{dst}$ during the training process to ensure the features aggregated with fewer neighboring frames can perform similarly to those aggregated with the whole neighboring frames, so that the detection accuracy of objects with small/medium sizes can be improved. Here we use deformable dynamic aggregation as an example. We calculate the mean square loss ($\mathcal{L}_{\text{MSE}}$) between $\Delta \bm{f}_i$ and $\Delta \bm{f}_i^d$ as:
\begin{equation}
    \mathcal{L}_{dst} = \mathcal{L}_{\text{MSE}} \left(\Delta \bm{f}_i, \Delta \bm{f}_i^d\right)
    \label{eq: dstLossVanilla}
\end{equation}
Inplace distillation loss is only applied during the training process; thus, it will not affect the inference speed.
\begin{table}[!tb]
    \centering
    \caption{Performance comparison with the recent state-of-the-art video object detection approaches on ImageNet VID validation set. }
    \begin{tabular}{l|l|l|l|l|l|l|l|l}
    \toprule
    & Methods & FPS & mAP & AP$_{0.5}$ &  AP$_{0.75}$ & AP$_s$ & AP$_m$ & AP$_l$\\
    \midrule
    \multirow{9}{*}{\rotatebox{90}{ResNet-50}} & FGFA \cite{zhu2017flow} & 5.8 & 46.7 & 74.3 & 51.5 & 5.7 & 21.8 & 52.9 \\
    & FGFA + Vanilla DA & 7.9 & 46.4 & 73.9 & 51.1 & 5.3 & 20.8 & 52.5\\
    & FGFA + Deformable DA & 7.6 & 46.6 & 74.1 & 51.2 & 6.5 & 21.6 & 52.7\\
    \cline{2-9}
    \rule{0pt}{10pt} & SELSA \cite{wu2019sequence} & 5.0 & 48.1 & 77.9 & 52.8 & 8.3 & 26.2 & 54.3\\
    & SELSA + Vanilla DA & 9.4 & 47.2 & 76.5 & 51.1 & 7.6 & 25.8 & 52.9 \\
    & SELSA + Deformable DA & 8.8 & 47.9 & 77.5 & 52.4 & 8.7 & 26.1 & 53.5\\
    \cline{2-9}
    \rule{0pt}{10pt} & Temporal ROI Align \cite{gong2021temporal} & 1.5 & 48.1 & 79.0 & 52.1 & 7.0 & 26.2 & 54.1\\
    & Temporal ROI Align + Vanilla DA & 3.9 & 46.9 & 77.8 & 51.0 & 6.4 & 25.7 & 52.6\\
    & Temporal ROI Align + Deformable DA & 3.5 & 47.8 & 78.8 & 51.7 & 7.2 & 25.9 & 53.5\\
    \midrule
    \multirow{9}{*}{\rotatebox{90}{ResNet-101}} & FGFA & 5.1 & 50.2 & 77.6 & 56.1 & 7.3 & 24.0 & 56.3\\
    & FGFA + Vanilla DA & 7.5 & 49.7 & 77.2 & 54.9 & 6.9 & 24.1 & 56.1\\
    & FGFA + Deformable DA & 7.1 & 50.1 & 77.5 & 55.8 & 7.9 & 23.8 & 56.1 \\
    \cline{2-9}
    \rule{0pt}{10pt} & SELSA \cite{wu2019sequence} & 4.5 & 52.1 & 81.3 & 57.4 & 9.0 & 28.1 & 58.1\\
    & SELSA + Vanilla DA & 8.5 & 51.2 & 80.0 & 57.0 & 7.8 & 26.8 & 57.3\\
    & SELSA + Deformable DA & 8.0 & 52.0 & 81.0 & 56.8 & 9.1 & 27.9 & 57.8 \\
    \cline{2-9}
    \rule{0pt}{10pt}&  Temporal ROI Align \cite{gong2021temporal} & 1.2 & 51.3 & 82.4 & 56.1 & 10.4 & 28.7 & 56.9\\
    & Temporal ROI Align + Vanilla DA & 3.6 & 50.4 & 81.8 & 55.3 & 9.3 & 27.5 & 55.1\\
    & Temporal ROI Align + Deformable DA & 3.3 & 50.9 & 82.0 & 55.6  & 10.5 & 29.5 & 56.3 \\
    \bottomrule
    \end{tabular}
    \label{tab: mainResult}
\end{table}
\section{Experiments}
\subsection{Experiment Setup.}
For mini-network $\mathcal{N}_{mot}^v$ and $\mathcal{N}_{mot}^d$, a one-layer convolutional layer is used to fuse the concatenated features. Then a global average pooling operation is applied to reduce the spatial and temporal resolutions. Next, the pooled feature is fed into a 2-layer MLP for classification $(\mathcal{N}_{mot}^v)$ or regression $(\mathcal{N}_{mot}^d)$. The object size estimation network $\mathcal{N}_{size}$ has the same architecture as $\mathcal{N}_{mot}^d$ except that the input is $\bm{f}_i$ rather than the concatenation of $\bm{f}_i$ and $\bm{f}_j$. For vanilla dynamic aggregation, $\theta$ is set to be $3$ unless otherwise stated. 

We evaluate our proposed methods on the ImageNet VID benchmark \cite{russakovsky2015imagenet} as the recent state-of-the-art video object detection models \cite{chen2020memory, wu2019sequence, zhu2017flow, deng2019relation}. Following the widely used protocols in \cite{zhu2017flow, chen2020memory, wu2019sequence}, we train our model on a combination of ImageNet VID and DET datasets. We implement our method mainly based on mmtracking\cite{mmtrack2020}\footnote{There are around $2\%$ mAP fluctuations in performance, and we take the mean after running $5$ experiments.}. The whole network is trained on 8 Tesla A100 GPUs. During the inference process, $30$ neighboring frames are used for feature aggregation.

\subsection{Main Results}
In this section, we conduct experiments on vanilla, and deformable feature aggregation with the current video object detectors on the ImageNet VID benchmark \cite{russakovsky2015imagenet}. We compare state-of-the-art feature aggregation-based video object detectors integrated with our proposed methods. The results are summarized in Table \ref{tab: mainResult}. For local aggregation methods like FGFA \cite{zhu2017flow}, our proposed dynamic aggregation can significantly improve the inference speeds while maintaining comparable performance like mAP and AP$_{0.5}$. We argue that this is because FGFA aggregates feature with local temporal neighboring frames, which share much redundant information. Therefore, removing those redundancies during the inference process will not affect the final predictions much, especially when the objects are at slow motion speeds.

\begin{figure}[!tb]
    \centering
    \includegraphics[width=10cm]{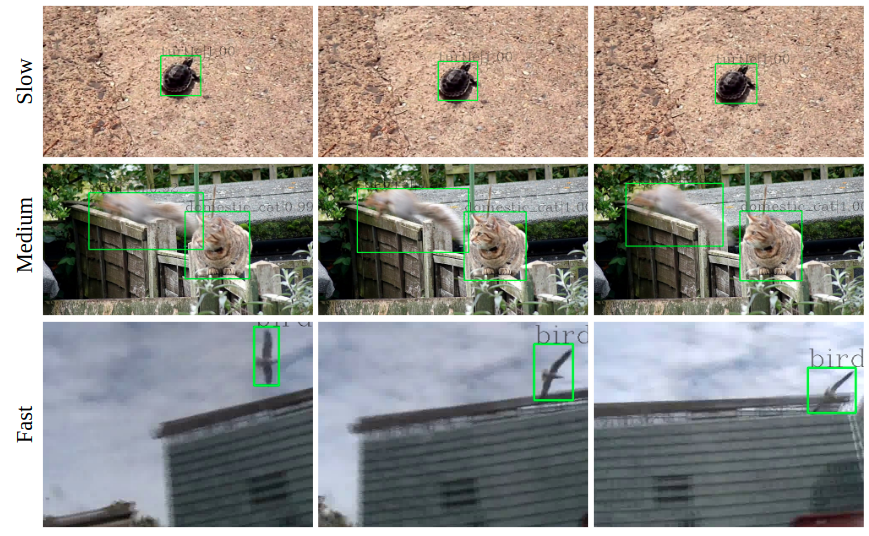}
    \caption{Examples of FGFA \cite{zhu2017flow} integrated with deformable dynamic aggregation on detecting objects with different motion speeds in the video.}
    \label{fig: vidEx}
\end{figure}

For global aggregation methods like SELSA \cite{wu2019sequence}, and Temporal ROI Align \cite{gong2021temporal}, our proposed methods can still improve the inference speeds by a large margin yet at the sacrifice of performance like AP$_{0.75}$. We argue that this is because global aggregation methods select features with similar representations for aggregation, and removing several frames during aggregation may have a harmful effect on precisely localizing the bounding boxes, considering AP$_{0.75}$ drops more compared with AP$_{0.5}$. Meanwhile, we notice that compared with vanilla dynamic aggregation, the deformable version has much better performance (even better than the original model) on small object detection when taking the object sizes into account, which validates the effectiveness of the proposed modules. We argue that our methods can adaptively select the frames for aggregation, which provide adequate but not redundant information for object detection. Figure \ref{fig: vidEx} shows several examples of video object detection results integrated with deformable dynamic aggregation. From the figure, our proposed methods can precisely predict the bounding boxes and categories of objects in each video frame. 

\subsection{Model Analysis}
In this section, we conduct extensive ablation study experiments to analyze the structures and parameters of our proposed modules. By default, we use FGFA \cite{zhu2017flow} with the backbone of ResNet-50 \cite{he2016deep} as the model to conduct experiments unless otherwise stated. In this section, we mainly analyze the proposed deformable dynamic aggregation module. 

\noindent\subsubsection{Analysis of sampling function $\mathcal{G}$.} We conduct experiments with deformable feature aggregation on the choices of sampling function as Table \ref{tab: selectionG}. ``Nearest'' and ``Furthest'' represent choosing the closest and furthest $k_d$ frames for aggregation, while ``Bin'' means binning the $k$ frames into $k_d$ buckets and sample $1$ frame from each bucket. For example, suppose the current frame is the $11^{th}$ of a video with $21$ frames and $k_d = 7$, Table \ref{tab: sampling} shows the comparison of selected frames with different sampling functions. Besides the three sampling functions mentioned above, we also compare with random sampling results. From Table \ref{tab: sampling}, ``Nearest'' sampling has the best performance compared with the other methods, while ``Bin'' sampling has a comparable result. ``Furthest'' and ``Random'' sampling methods have a poor performance, and we argue that this is because there are not enough effective and informative frames for aggregation when using these two strategies.
\begin{table}[!tb]
    \centering
    \caption{Comparison of video object detection results with different sampling functions $\mathcal{G}$ on FGFA \cite{zhu2017flow} with ResNet-50 \cite{he2016deep} as the backbone.}
    \begin{tabular}{l|l|l|l}
    \toprule
    Method &  Nearest & Furthest & Bin\\
    \midrule
    Selected Frames & 8, 9, 10, 11, 12, 13, 14 & 1, 2, 3, 11, 19, 20, 21 & 2, 5, 8, 11, 14, 17, 20\\
    \bottomrule
    \end{tabular}
    \label{tab: sampling}
\end{table}
\begin{table}[!tb]
    \centering
    \caption{Example of selected frames with different sampling function $\mathcal{G}$.}
    \begin{tabular}{l|l|l|l|l|l|l}
    \toprule
    Method & mAP & AP$_{0.5}$ & AP$_{0.75}$ & AP$_s$ & AP$_m$ & AP$_l$\\
    \midrule
    Nearest & 47.0 & 74.5 & 51.2 & 6.5 & 21.6 &  53.3\\
    Furthest & 45.5 & 72.5 & 50.1 & 5.6 & 20.8 & 51.7\\
    Random & 45.9 & 73.7 & 50.6 & 5.8 & 21.0 &  52.3\\
    Bin & 46.8 & 74.4 & 51.0 & 6.4 & 21.4 &  53.2\\
    \bottomrule
    \end{tabular}
    \label{tab: selectionG}
\end{table}

\noindent\subsubsection{Analysis of mapping function $\sigma$.} We analyze the choice of mapping function $\sigma$ in our proposed deformable feature aggregation as Table \ref{tab: sigma}. We compare four different mapping functions, namely, linear, square root, quadratic and learnable function by retraining the models with the corresponding $\sigma$. From Table \ref{tab: sigma}, compared with linear function, when choosing learnable networks as mapping function, the performance is the best at the sacrifice of inference speed. Square root and quadratic functions can balance the inference speed and accuracy by mapping $s$ into different distributions.
\begin{table}[!tb]
    \centering
      \caption{Comparison of video object detection results with different mapping functions $\sigma$ on FGFA \cite{zhu2017flow} with ResNet-50 \cite{he2016deep} as the backbone.}
    \begin{tabular}{l|l|l|l|l|l|l|l}
    \toprule
    Function & FPS & mAP & AP$_{0.5}$ & AP$_{0.75}$ & AP$_s$ & AP$_m$ & AP$_l$\\
    \midrule
    Linear $(y=1-x)$ & 7.6 & 46.6 & 74.1 & 51.2 & 6.5 & 21.6 & 52.7 \\
    Sqrt $(y=1-\sqrt{x})$ & 7.9 & 46.4 & 74.0 & 51.0 & 6.1 & 21.3 & 52.6\\
    Quadratic $(y=1-x^2)$ & 7.4 & 46.7 & 74.3 & 51.4 & 6.7 & 21.5 & 52.7\\
    Learnable $\left(y=\texttt{MLP}(x)\right)$ & 7.1 & 46.9 & 74.4 & 51.5 & 6.9 & 21.9 & 53.1\\
    \bottomrule
    \end{tabular}
    \label{tab: sigma}
\end{table}

\noindent\subsubsection{Comparison with knowledge distillation.} We also conduct experiments to compare with knowledge distillation results. We notice that FGFA aggregated with $15$ frames have a similar inference speed as our proposed methods. Therefore, we use an FGFA aggregated with $15$ frames to distill the knowledge from an FGFA aggregated with $30$ frames and compare the results with our proposed method in Table \ref{tab: knowledgeDistill}. From the table, the distillation-only method is not as good as our proposed methods despite similar inference speed. Also, the model will perform worse if trained without inplace distillation loss.
\begin{table}[!tb]
    \centering
     \caption{Comparison between FGFA distilled from a model aggregated with more frames and the proposed method. $\dagger$ means models without inplace distillation loss.}
    \begin{tabular}{l|l|l|l|l|l|l|l}
    \toprule
    Method & FPS & mAP & AP$_{0.5}$ & AP$_{0.75}$ & AP$_s$ & AP$_m$ & AP$_l$\\
    \midrule
    FGFA (15 frames) & 7.7 & 46.2 & 73.7 & 50.5 & 5.8 & 20.8 & 52.0\\
    FGFA (15 frames) + Distill & 7.7 & 46.4 & 73.9 & 50.7 & 5.9 & 21.0 & 52.3\\
    FGFA (30 frames) & 5.8 & 46.7 & 74.3 & 51.5 & 5.7 & 21.8 & 52.9\\
    FGFA (30 frames)+ Ours$^\dagger$ & 7.6 & 46.5 & 73.9 & 50.9 & 5.6 & 20.9 & 52.5\\
    FGFA (30 frames)+ Ours & 7.6 & 46.6 & 74.1 & 51.2 & 6.5 & 21.6 & 52.7\\
    \bottomrule
    \end{tabular}
    \label{tab: knowledgeDistill}
\end{table}

\noindent\subsubsection{Analysis of motion speeds.} Besides object sizes, we also compare experimental results to analyze the effects on object motion speeds. Following MEGA \cite{chen2020memory}, we categorize objects into slow, medium, and fast groups and calculate their corresponding accuracy as Table \ref{tab: motionSpeed}. The table shows that the performance on objects with slow motion speeds drops a little when integrated with a deformable dynamic aggregation module. However, detection accuracy on objects with medium motion speeds decreases by $0.4\%$. We argue that this is because situations like occlusion or rare positions are always treated as objects with medium motion speeds, and a few frames are sampled for aggregation, which are not enough to handle those cases.
\begin{table}[!tb]
    \centering
    \caption{Comparison between FGFA integrated with/without our proposed methods on video object detection on objects with different motion speeds.}
    \begin{tabular}{l|l|l|l|l|l|l}
    \toprule
    Method & FPS & mAP & AP$_{50}$& AP$_{\text{slow}}$ & AP$_{\text{medium}}$ & AP$_{\text{fast}}$\\
    \midrule
    FGFA & 5.8 & 46.8 & 74.3 & 83.8 & 72.2 & 50.5\\
    FGFA + Ours & 7.6 & 46.5 & 74.2 & 83.7 & 71.8 & 50.3\\
    \bottomrule
    \end{tabular}
    \label{tab: motionSpeed}
\end{table}
\section{Conclusion}
Existing feature aggregation-based video object detectors usually apply a fixed number of frames to enhance objects' representations and boost performance. Therefore, the performance and inference speed are heavily influenced by the number of frames used for aggregation. In this paper, we aim to perform dynamic aggregation to the current methods to balance the performance and inference speed. We first propose vanilla dynamic aggregation and then extend to a deformable version which can adaptively and reconfigurably select frames used for feature enhancement according to the input frames. Furthermore, we introduce the inplace distillation loss to boost the performance of frames not fully aggregated. Extensive experiments on the ImageNet VID benchmark validate the effectiveness and efficiency of our proposed methods. We hope our approaches can bring some ideas to the efficient video object detection field.

\bibliographystyle{splncs04}
\bibliography{ref}
\end{document}